%% file: main.tex
\documentclass[lettersize,journal]{IEEEtran}
\usepackage{amsmath,amsfonts}
\usepackage{algorithmic}
\usepackage{algorithm}
\usepackage{array}
\usepackage[caption=false,font=normalsize,labelfont=sf,textfont=sf]{subfig}
\usepackage{textcomp}
\usepackage{stfloats}
\usepackage{url}
\usepackage{verbatim}
\usepackage{graphicx}
\usepackage{cite}
\begin{document}

\title{Multi-task GINN-LP for Multi-target Symbolic Regression}

\author{\IEEEauthorblockN{Hussein Rajabu, Lijun Qian, and Xishuang Dong} \\
\IEEEauthorblockA{Department of Electrical and Computer Engineering \\ 
Prairie View A\&M University, 
Prairie View, TX 77446, USA \\
Email: hrajabu@pvamu.edu, liqian@pvamu.edu xidong@pvamu.edu}

\thanks{This work has been submitted to the IEEE for possible publication. Copyright may be transferred without notice, after which this version may no longer be accessible.}
}

\markboth{Journal of \LaTeX\ Class Files,~Vol.~14, No.~8, August~2021}%
{Shell \MakeLowercase{\textit{et al.}}: A Sample Article Using IEEEtran.cls for IEEE Journals}


\maketitle

\begin{abstract}

In the area of explainable artificial intelligence, Symbolic Regression (SR) has emerged as a promising approach by discovering interpretable mathematical expressions that fit data. However, SR faces two main challenges: most methods are evaluated on scientific datasets with well-understood relationships, limiting generalization, and SR primarily targets single-output regression, whereas many real-world problems involve multi-target outputs with interdependent variables. To address these issues, we propose multi-task regression GINN-LP (MTRGINN-LP), an interpretable neural network for multi-target symbolic regression.  By integrating GINN-LP with a multi-task deep learning, the model combines a shared backbone including multiple power-term approximator blocks with task-specific output layers, capturing inter-target dependencies while preserving interpretability. We validate multi-task GINN-LP on practical multi-target applications, including energy efficiency prediction and sustainable agriculture. Experimental results demonstrate competitive predictive performance alongside high interpretability, effectively extending symbolic regression to broader real-world multi-output tasks.

\end{abstract}

\begin{IEEEkeywords} Interpretable AI, Multi-task Learning, Multi-target Regression, Symbolic AI; 
\end {IEEEkeywords}

\section{Introduction}
\label{sec1}
\input{Introduction}

\section{Task Definition}
\label{sec2}
\input{Task}

\section{Methodology}
\label{sec3}
\input{Method}

\section{Experiments}
\label{sec4}
\input{Experiment}

\section{Related Work}
\label{sec5}
\input{Relatedwork}
\section{Conclusion and Future Work}
\label{sec6}
\input{Conclusion}

\section*{Acknowledgment}
This research work is partially supported by the U.S. NSF under award number 2235731, 2323419, 24018601 and by the Army Research Office (ARO) under cooperative agreement W911NF-24-2-0133. The views and conclusions contained in this document are those of the authors and should not be interpreted as representing the official policies, either expressed or implied, of NSF or ARO or the U.S. Government. The U.S. Government is authorized to reproduce and distribute reprints for Government purposes notwithstanding any copyright notation herein. Additionally, the authors acknowledge the use of AI-based tools, such as ChatGPT, for assistance in editing, grammar enhancement, and spelling checks during the preparation of this manuscript.



\bibliographystyle{IEEEtran}
\bibliography{References}




 \begin{IEEEbiography}[{\includegraphics[width=1in,height=1.25in,clip,keepaspectratio]{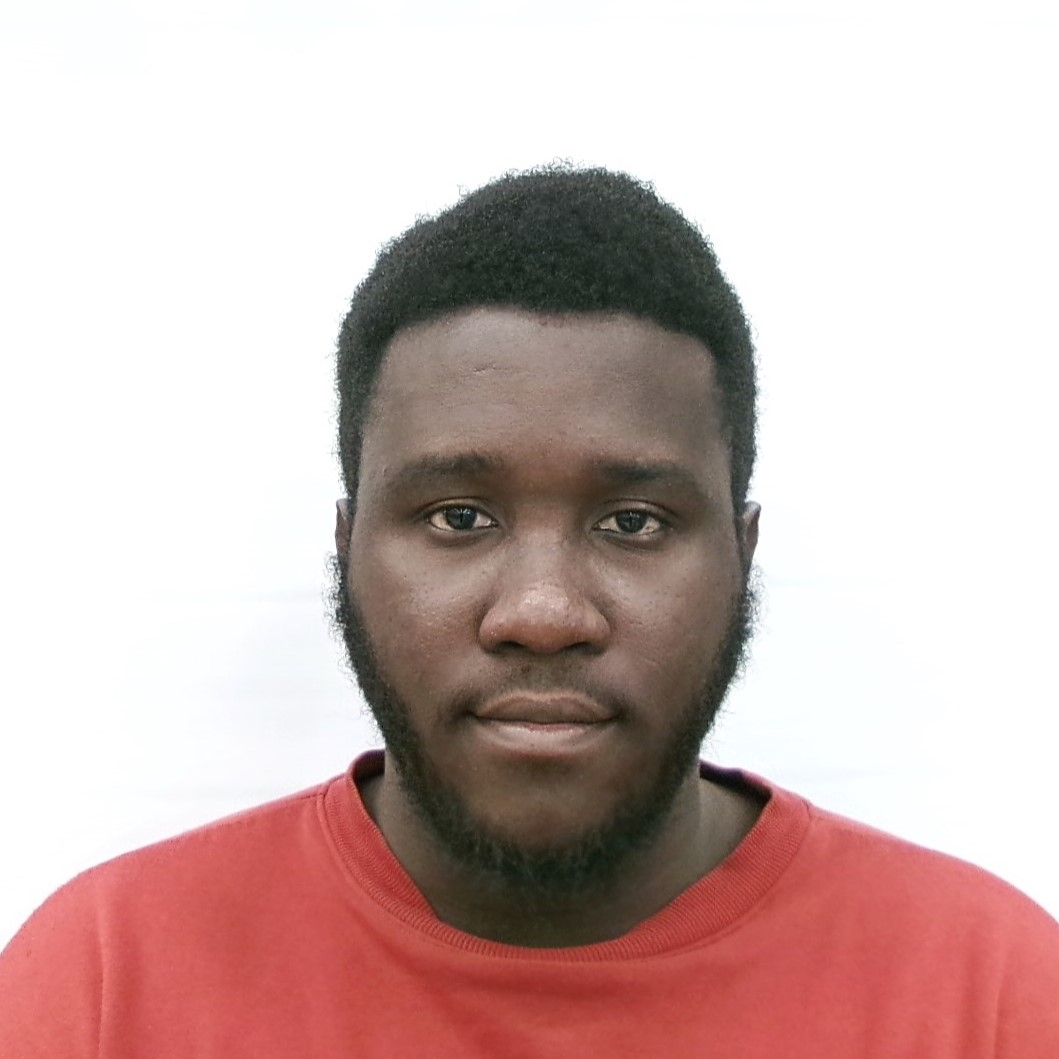}}] {Hussein Rajabu} received the B.S. degree from University of Houston, Houston, USA. He is currently a M.S. Student and a Graduate Assistant Researcher with the Department of Electrical and Computer Engineering, Prairie View A\&M University (PVAMU), Prairie View, TX, USA. His research interests are in the areas of interpretable AI and Uncertainty Awareness 3D Reconstruction.
 
\end{IEEEbiography}

\vspace{11pt}

 \begin{IEEEbiography}[{\includegraphics[width=1in,height=1.25in,clip,keepaspectratio]{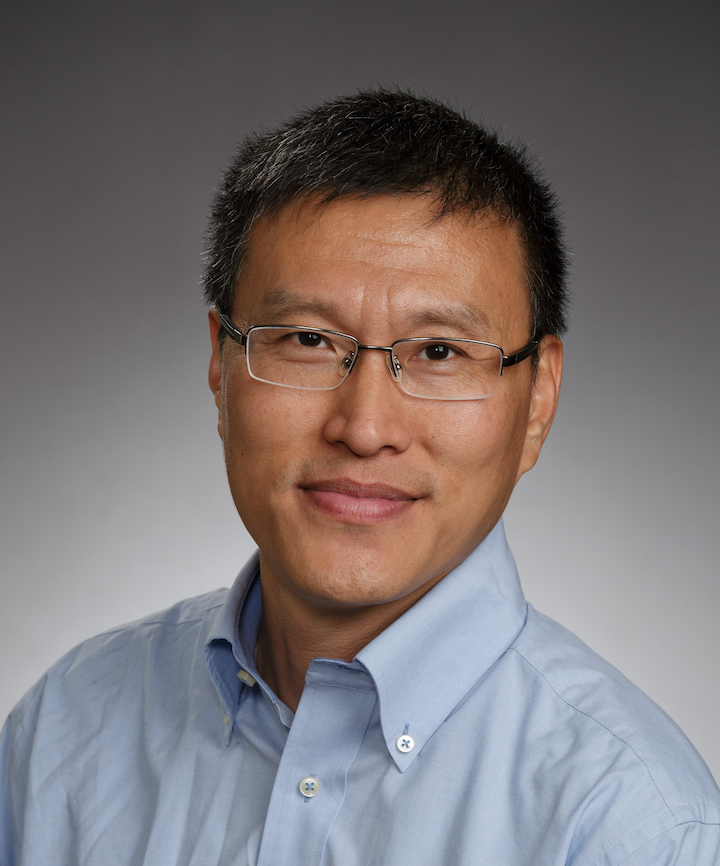}}]{Lijun Qian} received the B.S. degree from Tsinghua University, Beijing, China, the M.S. degree from the Technion - Israel Institute of Technology, Haifa, Israel, and the Ph.D. degree from Rutgers University, USA.

He is currently a Regents Professor and holds the AT\&T Endowment with the Department of Electrical and Computer Engineering, Prairie View A\&M University (PVAMU), Prairie View, TX, USA. He is also the founder and director of the Center of Excellence in Research and Education for Big Military Data Intelligence (CREDIT Center). 

Dr. Qian was a Member of technical staff of Bell-Labs Research, Murray Hill, NJ, USA. He was a Visiting Professor with Aalto University, Finland. His research interests are in the areas of artificial intelligence, machine learning, big data analytics, wireless communications and mobile networks, network security, and computational systems biology.
\end{IEEEbiography}

\vspace{11pt}

\begin{IEEEbiography}[{\includegraphics[width=1in,height=1.25in,clip,keepaspectratio]{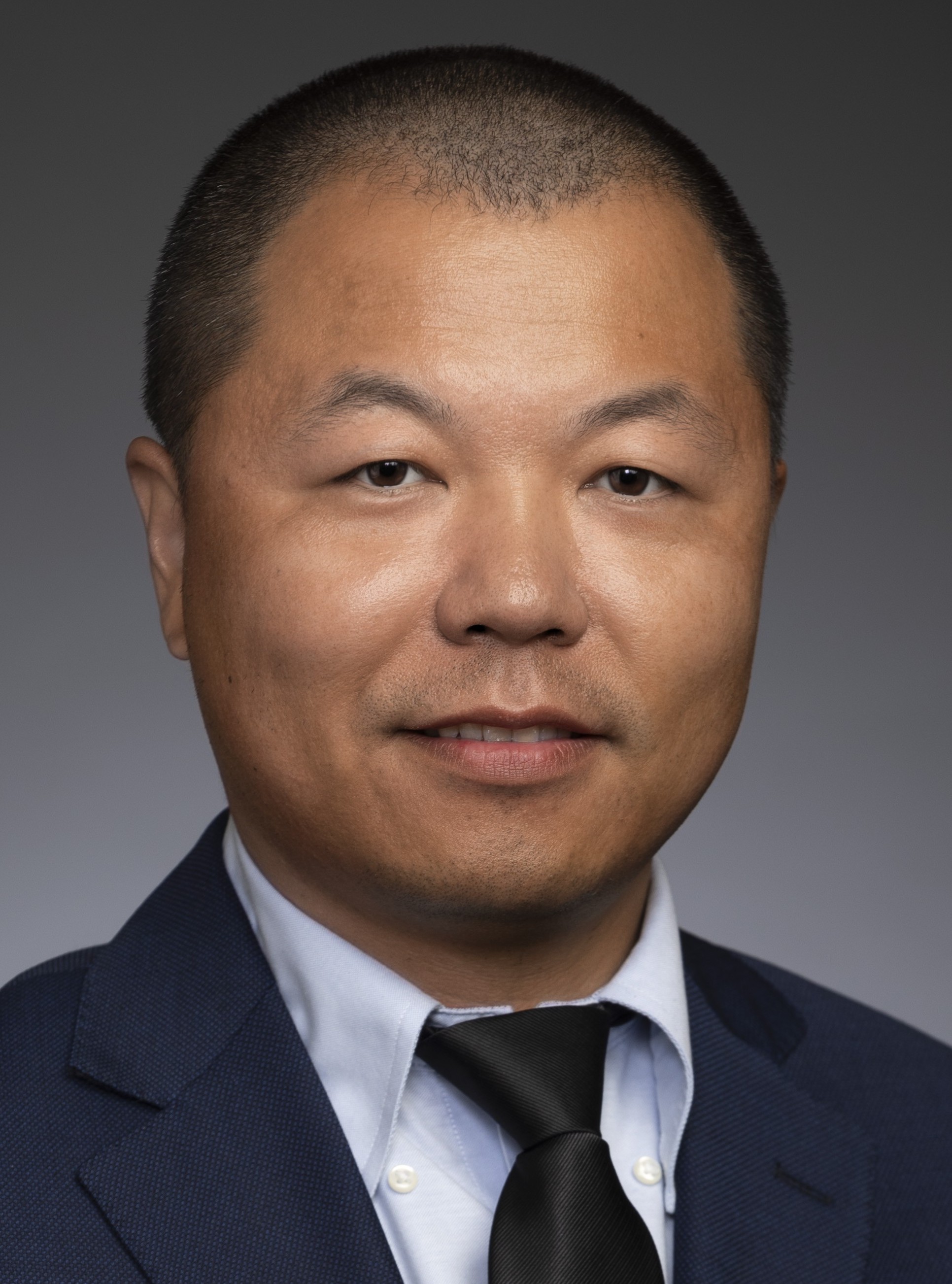}}]{Xishuang Dong} received the B.S. degree from Harbin University of Science and Technology, Harbin, China, the M.S. degree from Harbin Engineering University, Harbin, China, and the Ph.D. degree from Harbin Institute of Technology University, Harbin.

He is currently an Associate Professor with the Department of Electrical and Computer Engineering, Prairie View A\&M University (PVAMU), Prairie View, TX, USA. His research interests include generative artificial intelligence, deep learning, digital twins, and computational systems biology.
\end{IEEEbiography}

\vfill

\end{document}

%% file: Introduction.tex

\begin{figure*}[!h]
    \centering
    \includegraphics[width=0.8\linewidth]{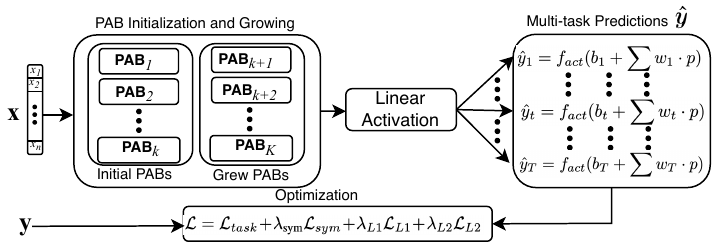}
    \caption{Diagram of the multi-task regression GINN-LP (MTRGINN-LP). Multiple Polynomial Approximation Blocks (PABs) are initialized and progressively expanded during training. The model then produces multi-task predictions $\hat{y}$ by applying linear activation functions $f_{\text{act}}(\cdot)$ to independently weighted sums of polynomial terms from the PABs. Finally, based on the predictions $\hat{y}$ and ground truth $\mathbf{y}$, the model is optimized by minimizing the total loss $\mathcal{L}$.}    \label{fig_mtginn-lp}
\end{figure*}

Over the past decade, deep learning has achieved remarkable success across diverse domains such as computer vision~\cite{chai2021deep} and natural language processing, largely due to its expansive hypothesis space enabled by deep signal propagation through multiple hidden layers~\cite{pouyanfar2018survey, ji2025comprehensive}. Despite their remarkable predictive capabilities, without a clear understanding of how these models reach their predictions, it is extremely challenging to build user trustable applications, especially in high-stakes decision-making contexts~\cite{zhang2024neuro}.

To address this issue, explainable artificial intelligence (XAI) has gained increasing attention in recent years, aiming to bridge the gap between predictive performance and human understanding~\cite{guidotti2018survey}. Model-agnostic methods such as Local Interpretable Model-agnostic Explanations (LIME) approximate local decision boundaries using interpretable surrogate models~\cite{ribeiro2016should}, while SHapley Additive exPlanations (SHAP) assign feature importance scores based on Shapley values~\cite{lundberg2017unified}. Although these post-hoc explanation techniques are invaluable for enhancing human understanding, they primarily describe statistical associations between inputs and outputs rather than uncovering the underlying mechanisms by which models generate their predictions~\cite{somvanshi2025bridging}.

In recent years, symbolic regression (SR) has attracted significant attention in the field of XAI~\cite{makke2024interpretable}. SR is an optimization problem that seeks the most suitable mathematical expression or model to fit a given dataset~\cite{koza1994genetic}, aiming not only for high predictive accuracy but also for interpretability. A model is considered interpretable if the relationship between its inputs and outputs can be logically or mathematically traced in a concise manner. In other words, models are interpretable when they can be represented as explicit mathematical equations. Over the past decade, SR has undergone rapid development due to its inherent interpretability and wide applicability, leading to a growing number of algorithms and practical implementations.

However, two major challenges remain~\cite{wang2019symbolic, angelis2023artificial, dong2025recent, aldeia2025call}: 1) Most existing SR methods have been developed and evaluated primarily on scientific datasets, particularly in physics and chemistry, where the underlying relationships between input and output  are often well understood. Nevertheless, limited validation on broader machine learning tasks constrains its generalization and applicability to more diverse domains; and 2) SR research has largely focused on single-target regression, where the goal is to derive an explicit mathematical expression mapping input variables to a single continuous output. In contrast, many real-world problems involve multi-target outputs with interdependent variables or shared latent relationships.

To address these challenges, we propose multi-task regression GINN-LP (MTRGINN-LP)~\cite{ranasinghe2024ginn}, a neuro-symbolic AI framework designed to perform multi-target symbolic regression. GINN-LP is an interpretable neural network capable of discovering both the structure and coefficients of an underlying equation, assuming the equation can be represented as a multivariate Laurent Polynomial. The core of GINN-LP is an interpretable neural network component called the power-term approximator block (PAB), which leverages logarithmic and exponential activation functions to capture multiplicative and power-law relationships. We extend GINN-LP to the multi-target setting by integrating it with a multi-task deep learning framework~\cite{crawshaw2020multi}. Specifically, the model employs a shared backbone composed of multiple PABs, followed by task-specific output layers, each formed as a linear combination of the shared PAB outputs. This design enables the model to capture shared representations across related targets while preserving task-specific interpretability. Additionally, we incorporate a symbolic loss that reduces the discrepancy between symbolic predictions and regression targets during training.

Furthermore, the proposed approach is validated on broader machine learning tasks beyond traditional scientific domains, including energy efficiency prediction and sustainable agriculture, both focusing on multi-target regression problems. Experimental results demonstrate that the proposed method achieves competitive predictive performance while maintaining high interpretability, effectively mitigating the gap between symbolic regression and practical multi-output applications.

The contributions of this study are summarized as follows:

\begin{itemize}
\item We extend the interpretable GINN-LP architecture to support multi-target regression by integrating it with the multi-task deep learning. The model employs a shared backbone of PABs with task-specific linear output layers, enabling shared representation learning while preserving interpretability. 

\item We introduce a symbolic loss that minimizes the difference between symbolic predictions and regression targets during training. It enhances both the performance of symbolic predictions and the overall model performance simultaneously.

\item We evaluate the proposed multi-task GINN-LP on diverse domains beyond traditional scientific applications, including energy efficiency prediction and sustainable agriculture. Experimental results show that it achieves competitive performance while maintaining transparent, interpretable representations, bridging the gap between symbolic regression and practical multi-output modeling.
 
\end{itemize}

%% file: Task.tex

Traditional symbolic regression aims to learn mathematical expressions from data that capture generalizable  patterns for downstream applications. Given a dataset \( D = \{(\mathbf{x_i}, y_i)\}_{i=1}^n \), where \( \mathbf{x_i} \in \mathbb{R}^d \) denotes the input vector and \( y_i \in \mathbb{R} \) represents a scalar output. Let \( f \) be a function expressed symbolically that defines a mapping \( f : \mathbb{R}^d \rightarrow \mathbb{R} \). The loss function is defined as below.

\begin{equation}
\mathcal{L}_{SR}(f)= \sum_{i=1}^n \ell (f(\mathbf{x_i}), y_i)
\end{equation}

where the mean squared error (MSE) can be used as the loss function $\ell(\cdot)$. 

This study extends the traditional symbolic regression to multi-target symbolic regression (MSR) problem as following.

\begin{equation}
\min\mathcal{L}_{MSR}(f) = \frac{1}{M} \sum_{j=1}^{M} \sum_{i=1}^{N} \ell_{j} \left(y_i^{(j)}, f_j(\mathbf{x}_i)\right)
\end{equation}

where 

\begin{itemize}
\item \( D = \{(\mathbf{x_i}, \mathbf{y_i})\}_{i=1}^n \) with \( \mathbf{x_i} \in \mathbb{R}^d \) as the input vector and \( \mathbf{y_i} \in \mathbb{R^M} \)  as the multi-dimensional  vector containing $M$ targets. Specifically, $y_i^{(j)}$ refers to the $j$-th target of  $i$-th sample.

\item $\ell_{j}(\cdot)$ refers to the loss for each target, typically the MSE.

\end{itemize}

%% file: Method.tex

\subsection{Multi-task Learning}

Multi-Task Learning (MTL) is a machine learning paradigm that seeks to exploit useful information shared across multiple related tasks to enhance the generalization performance of all tasks~\cite{zhang2021survey}.  Let the task set be defined as \( \mathcal{T} = \{ \tau_1, \tau_2, \ldots, \tau_t, \ldots, \tau_n \} \) with \( n \) tasks. For each task \( \tau_t \), the input space, output space, and dataset are denoted by \( \mathcal{X} \), \( \mathcal{Y}_t \), and \( D_t = \{ (x_i^t, y_i^t) \}_{i = 1}^{N_t} \), respectively. All tasks share a common input space \( \mathcal{X} \).  MTL has been successfully applied across diverse domains, including speech recognition~\cite{deng2013new}, drug discovery~\cite{ramsundar2015massively}, computer vision~\cite{girshick2015fast}, and natural language processing~\cite{collobert2008unified}.

Multi-Task Deep Learning (MTDL) extends MTL by employing deep neural networks~\cite{ruder2017overview}, allowing shared and hierarchical representations to be learned directly from raw data. Unlike conventional MTL, which often relies on manual feature engineering or shallow architectures with explicit parameter sharing, MTDL leverages deep feature extractors that are jointly trained across tasks. Each task maintains its own output layer, while the shared deep backbone enables end-to-end learning on complex, high-dimensional data such as images, text, and sequential inputs.

\subsection{Growing Interpretable Neural Network - Laurent Polynomial (GINN-LP)}

Growing Interpretable Neural Network–Laurent Polynomial (GINN-LP)~\cite{ranasinghe2024ginn} is an interpretable neural architecture that uncovers both the functional form and coefficients of an underlying equation, assuming a multivariate Laurent polynomial relationship. It introduces the power-term approximator block (PAB), which uses logarithmic and exponential activations to approximate power terms, and employs a neural growth strategy to automatically determine the optimal number of PABs. Starting with a single PAB, the model incrementally adds new, randomly initialized blocks while retaining previously trained parameters to prevent overfitting. This iterative expansion continues until an early-stopping condition or maximum network size is reached, after which the final output is represented as a linear combination of all PABs.

\subsection{Proposed Method}

This study proposes MTRGINN-LP  to address symbolic multi-task regression problems. Figure~\ref{fig_mtginn-lp} illustrates the framework of the proposed method, which begins by initializing multiple PABs. Each PAB is formulated under the assumption that the underlying mathematical relationship in the data follows the structure of a multivariate Laurent polynomial.

\begin{align}
f(\mathbf{x}) = \sum_{i = 1}^{n} w_{i}log(x_{i}) = log(\prod_{i=1}^{n}x_{i}^{w_{i}})  
\end{align}

\begin{align}
p = e^{f(\mathbf{x})} = e^{\sum_{i = 1}^{n} w_{i}log(x_{i})}  = \prod_{i=1}^{n}x_{i}^{w_{i}}
\label{eq_pab}
\end{align}

\begin{algorithm}[t]
\caption{MTRGINN-LP Learning}
\label{alg:ginn_l_mt}
\textbf{Input:} Dataset $(\mathbf{x}, \mathbf{y})$, total epochs $E$. \\
\textbf{Output:} Symbolic equations $\{\hat{y}_{\text{sym}, 1}, \dots, \hat{y}_{\text{sym}, T}\}$. \\[3pt]

\textbf{For} epoch = 1 to $E$ \textbf{do} \\
\hspace*{1em}1. Compute model prediction $\hat{y}$ \\
\hspace*{1em}2. Compute task loss $\mathcal{L}_{task}$ \\
\hspace*{1em}3. Compute monomial features $m(\mathbf{x})$ \\
\hspace*{1em}4. Compute symbolic prediction $\hat{y}_{\text{sym}} $ \\
\hspace*{1em}5. Compute consistency loss: $\mathcal{L}_{sym} $ \\
\hspace*{1em}6. Compute total loss: \\
\hspace*{3em}$\mathcal{L} = \mathcal{L}_{task} + \lambda_{\text{sym}}\mathcal{L}_{sym} 
+ \lambda_{L1}\mathcal{L}_{L1} + \lambda_{L2}\mathcal{L}_{L2}$ \\
\hspace*{1em}7. Updating parameters $\theta$. \\
\hspace*{1em}8. Growing the model at each interval $G$.\\
\textbf{End For} \\[3pt]

\textbf{Return:} Symbolic equations 
$\hat{y}_{\text{sym}, t} = b_t + \sum_k w_{t,k}  \prod_i x_i^{w_{k,i}}$ for each task $t$.

\label{alg_mtginn-lp}
\end{algorithm}

Afterward, the model adopts a periodic growth strategy that automatically increases the number of PABs at predefined epoch intervals while retaining previously trained parameters of PABs. This strategy mitigates potential overfitting from an insufficient number of PABs during initialization for model learning. The iterative learning process continues until an early-stopping condition or maximum size of PABs is reached, and the final output $\hat{y}_t$ for target $t$ is expressed as a linear combination of all PABs, as formulated below.

\begin{align}
\hat{y}_t = f_{act}(b + \sum w \cdot p) =  f_{act}(b_t + \sum_k w_{t,k}  \prod_i x_i^{w_{k,i}})
\end{align}

where $i$, $k$, and $t$ indicate the $n$-th feature of the input $\mathbf{x}$, $n$-th PAB, and $t$-th task, respectively. $f_{act}(\cdot)$ denotes the activation function. The detailed learning process is outlined as algorithm~\ref{alg_mtginn-lp}. At each epoch, the algorithm first computes the neural network predictions $\hat{y}$ for all tasks. It then evaluates the task loss $\mathcal{L}_{task}$, which measures the difference between predicted outputs $\hat{y}$ and the true targets $\mathbf{y}$. For regression tasks, this is typically calculated as the mean squared error (MSE) between $\mathbf{y}$ and $\hat{y}$.

Next, the algorithm generates monomial features $m(\mathbf{x})$ from the input features, which serve as candidate terms for symbolic regression. Monomial features are terms formed by taking products of powers of input variables. Mathematically, a monomial in $n$ input features $\mathbf{x} = [x_1, x_2, \dots, x_n]$ can be written as:

\begin{align}
m(\mathbf{x}) = \prod_{i} x_i^{w_{k,i}}
\end{align}

where $a_i$ is a non-negative value, and $m(\mathbf{x})$ represents a single monomial feature. Each monomial is essentially a single feature transformation that combines input variables in a multiplicative way. In terms of $m(\mathbf{x})$, the symbolic prediction $\hat{y}_{\text{sym}, t} $ for target $t$ is calculated

\begin{align}
\hat{y}_{\text{sym}, t} = b + \sum w \cdot m(\mathbf{x})  =  b_t + \sum_k w_{t,k} \cdot m(\mathbf{x})
\end{align}

Subsequently, the symbolic predictions are then compared with the neural network predictions to compute the consistency loss $\mathcal{L}_{sym}$ with the mean squared error (MSE) between $\mathbf{y}_t$ and $\hat{y}_{\text{sym}, t}$ for target $t$. In addition, it employs $\lambda_{L1}\mathcal{L}_{L1}$ and $ \lambda_{L2}\mathcal{L}_{L2}$ to reduce model complexity and improve generalization. The output includes symbolic expressions $\hat{y}_{\text{sym}, t}$ for each task.

%% file: Experiment.tex

\subsection{Dataset}

This study employs two real-world datasets to validate the proposed method: Energy Efficiency\footnote{\url{https://archive.ics.uci.edu/dataset/242/energy+efficiency}} and Sustainable Agriculture\footnote{\url{https://www.kaggle.com/datasets/suvroo/ai-for-sustainable-agriculture-dataset}}. The Energy Efficiency dataset aims to evaluate the heating and cooling load requirements of buildings, i.e., their overall energy performance, as a function of various building parameters. It comprises 768 samples with 8 features and two continuous targets: Heating Load and Cooling Load. The Sustainable Agriculture dataset focuses on optimizing farming practices while promoting environmental and economic sustainability. It integrates information from farmers, weather stations, and market trends to support AI-driven, resource-efficient agricultural decision-making. This study employs 2,501 samples  with 15 features and two output targets: Sustainability Score and Consumer Trend Index, which respectively reflect the sustainability of farming practices and the market trend of agricultural products.    In summary, both datasets serve as benchmarks for validating the proposed method in solving two-task regression problems.

\subsection{Evaluation Metrics}

This study employs three standard evaluation metrics to evaluate the performance for regression problems, namely Mean Absolute Error (MAE), Mean Absolute Percentage Error (MAPE), and Root Mean Squared Error (RMSE)). Let $y_i$ denote the true value and $\hat{y}_i$ the predicted value for sample $i$, with $N$ total samples.  The MAE measures the average magnitude of prediction errors:

\begin{equation}
\text{MAE} = \frac{1}{N} \sum_{i=1}^{N} \left| y_i - \hat{y}_i \right|
\end{equation}

The MAPE measures the average absolute percentage error:

\begin{equation}
\text{MAPE} = \frac{100\%}{N} \sum_{i=1}^{N} \left| \frac{y_i - \hat{y}_i}{y_i} \right|
\end{equation}

The RMSE is defined as the square root of the mean squared error:

\begin{equation}
\text{RMSE} = \sqrt{\frac{1}{N} \sum_{i=1}^{N} \left( y_i - \hat{y}_i \right)^2 }
\end{equation}

MAE, RMSE, and MAPE are complementary metrics for evaluating regression performance. MAE measures the average absolute magnitude of errors and treats all deviations equally, making it easy to interpret and robust to outliers. RMSE, as the square root of the mean squared error, penalizes larger errors more heavily, highlighting models’ sensitivity to extreme deviations. MAPE expresses errors as percentages relative to the true values, allowing scale-free comparison across datasets, though it can be unstable when true values are near zero. Together, these metrics provide a balanced view: MAE captures general error magnitude, RMSE emphasizes significant deviations, and MAPE conveys relative predictive accuracy.

\begin{table}[h!]
\centering
\caption{Experiment setup for MTGINN-LP}
\begin{tabular}{l|c}
\hline
\textbf{Hyperparameter} & \textbf{Value} \\
\hline
Epoch & 2000 \\
\hline
Learning rate & 0.1 \\
\hline
Maximum PABs & 2, 4, 6, 8, 10 \\
\hline
Growing interval & 500 \\
\hline
$\lambda_{\text{sym}}$,  $\lambda_{L1}$, $\lambda_{L2}$ & 1e-2, 1e-4, 1e-4 \\
\hline
\end{tabular}

\label{tab-setup}
\end{table}

\begin{table*}[h!]
\centering
\setlength{\tabcolsep}{2pt} 
\caption{Performance comparison on predicting two targets: Heating Load ($Y_1$) and Cooling Load ($Y_2$), on the Energy Efficiency dataset. \textbf{Avg.} indicates the average across the two target variables. For example, \textbf{Avg. MAE} represents the average of \textbf{MAE (Y1)} and \textbf{MAE (Y2)}. MTRGINN-LP (Eq.) refers to the symbolic functions learned through training MTRGINN-LP.}
\begin{tabular}{lccccccccc}
\hline
\textbf{Model} & \textbf{MAE (Y1)} & \textbf{MAE (Y2)} & \textbf{Avg. MAE} & \textbf{MAPE (Y1)} & \textbf{MAPE (Y2)} & \textbf{Avg. MAPE} & \textbf{RMSE (Y1)} & \textbf{RMSE (Y2)} & \textbf{Avg. RMSE} \\
\hline
Linear Model   			& 2.23 & 2.43 & 2.33  & 10.25 & 9.39 & 9.82  & 3.14 & 3.49 & 3.31 \\
GP             			& 0.34 & 0.86 & 0.6  	  & 1.72 & 3.45 &  2.58 & 0.47 & 1.25 & 0.86  \\
MLP            			& 0.96 & 1.29 & 1.12  & 4.08 & 4.66 & 4.37  & 1.38 & 1.94 & 1.66 \\
Random Forest  		& 1.71 & 2.12 & 1.91  & 8.70 & 8.32 & 8.51 & 2.30 & 2.81 & 2.55 \\
SVR            			& 2.01 & 2.27 & 2.14 & 9.83 & 8.74 &  9.28 & 2.96 & 3.41 &  3.18 \\
\hline
\hline
\textbf{MTRGINN-LP}            	 		& 0.40 & 1.02 & 0.71  & 1.94 & 3.53 & 2.74  & 0.50 & 1.61 & 1.06  \\
\textbf{MTRGINN-LP (Eq.)}             	& 0.75 & 1.29 &  1.02 & 4.10 & 4.80 & 4.45 & 1.04 & 1.80 & 1.42 \\
\hline
\end{tabular}
\label{tab-enb}
\end{table*}

\subsection{Experiment Setup}

This study employs multiple traditional regression machine learning models as baselines to evaluate the effectiveness of the proposed method. The models include Random Forest, Linear Regression, Support Vector Regression (SVR), Multi-Layer Perceptron (MLP), and Gaussian Process (GP), all implemented using the scikit-learn library\footnote{https://scikit-learn.org/stable/}

The hyperparameters used to train the proposed methods are summarized in Table~\ref{tab-setup}.

\subsection{Results, Discussions, and Limitations}

This section presents the experimental results along with a discussion of the key observations and insights. It also summarizes the limitations of the proposed model.

\begin{figure*}[htbp]
    \centering
    \includegraphics[width=\linewidth]{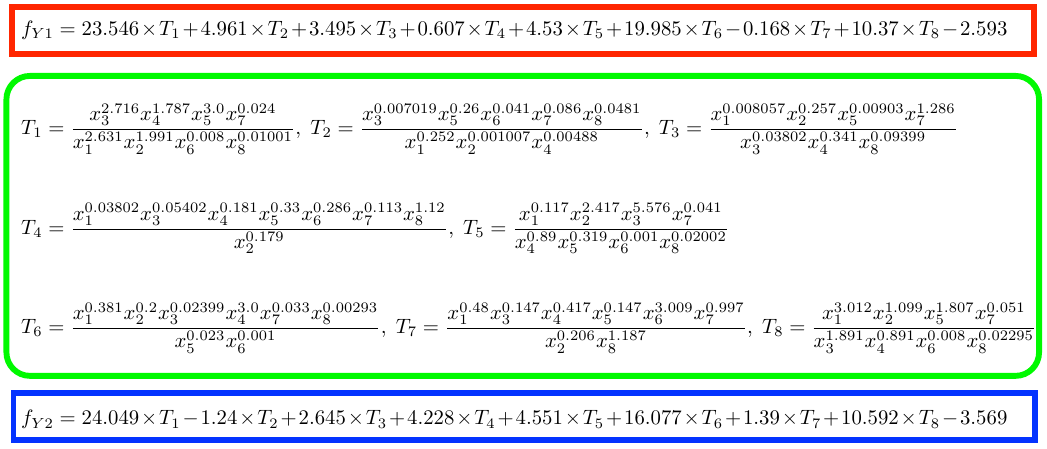}

    \caption{Symbolic functions learned by the proposal model for the Energy Efficiency datasets. For each target, the function $f(\cdot)$ consists of  eight terms ($T_1$ to $T_8$) and one bias term, corresponding to the optimal number 8 of PABs for the proposed model, where $f_{Y_1}$ and $f_{Y_2}$ are for estimating two targets: Heating Load and Cooling Load, respectively, respectively. Each term is a combination of eight input attributes ($x_1$ to $x_8$) with exponential form. }
    \label{fig:enb}
\end{figure*}

\subsubsection{Energy Efficiency}

Table~\ref{tab-enb} presents that traditional regression models such as Linear Regression, SVR, and Random Forest perform moderately on the Energy Efficiency dataset, with average MAE values ranging from 1.12 (MLP) to 2.33 (Linear Model). Gaussian Process (GP) achieves the lowest errors among these baselines, with an Avg. MAE of 0.60, demonstrating its strong predictive capability for this dataset. However, both MTRGINN-LP variants outperform most baselines in key metrics, particularly MTRGINN-LP, which achieves an Avg. MAE of 0.71 and an Avg. RMSE of 1.06. This indicates that the multi-target learning with the Laurent polynomial formulation effectively captures relationships across both targets, Y1 and Y2, while maintaining competitive accuracy and generalization.

\begin{figure*}[htbp]
    \centering
    \includegraphics[width=\linewidth]{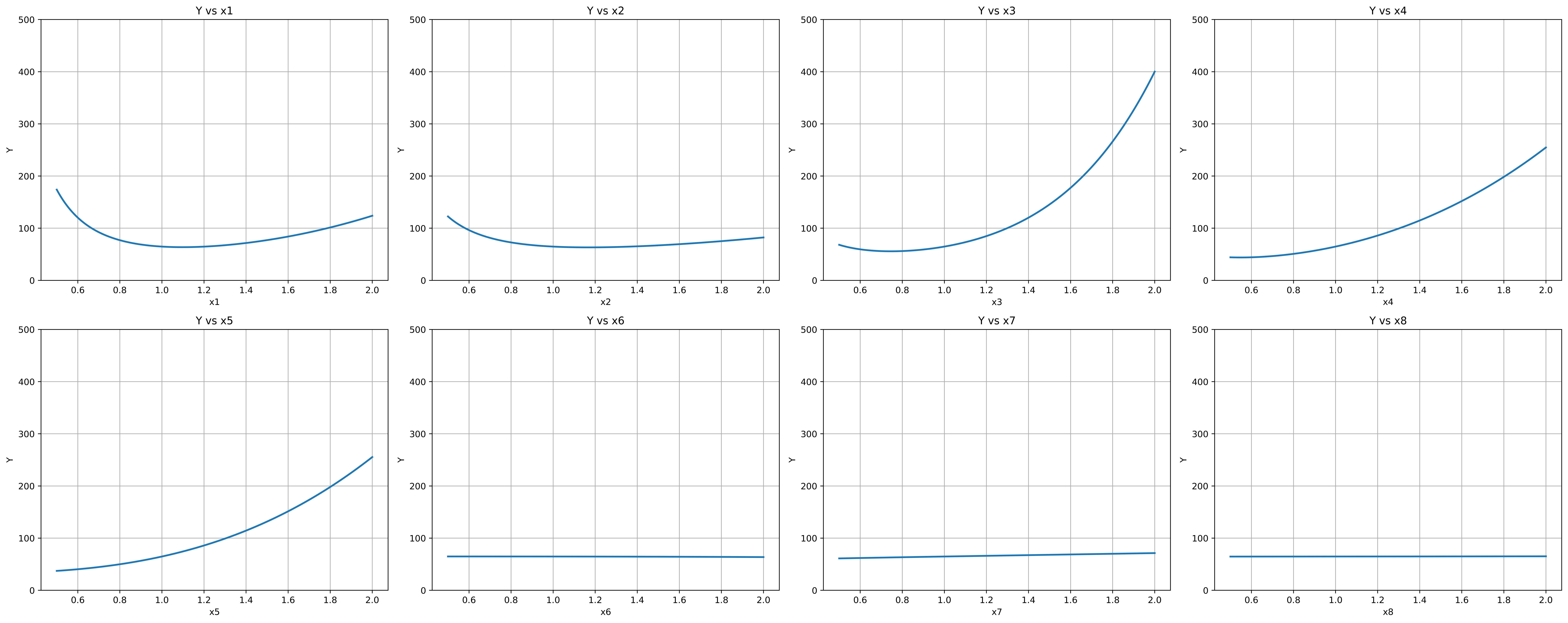}

    \caption{An example of correlation analysis between inputs and outputs in the symbolic functions. It is assumed that the input attributes are independent of one another, and the output includes only the target Heating Load denoted as $Y$.}
    \label{fig:enb_c}
\end{figure*}

While MTRGINN-LP (Eq.), the symbolic function extracted from the trained model, shows slightly higher errors (Avg. MAE of 1.02 and Avg. RMSE of 1.42) compared to the full model, it still outperforms most traditional baselines such as Linear Regression, SVR, and Random Forest. This suggests that the learned symbolic representations retain significant predictive power while providing interpretability. Overall, these results highlight the strength of the MTRGINN-LP framework in balancing predictive accuracy with symbolic transparency, making it particularly suitable for applications where both performance and explainability are important.

In addition, Figure~\ref{fig:enb} presents the symbolic functions learned by the proposal model. The model expresses the target outputs as linear combinations of eight symbolic terms and one bias term.  Each symbolic term is a nonlinear expression composed of eight input attributes in exponential form, capturing complex, interpretable relationships among the input variables. The optimal configuration of eight PABs ensures a balance between model complexity and accuracy in representing the underlying physical processes. The learned symbolic equations reveal that both targets share a similar structural form but differ in the magnitude and direction of the coefficients, reflecting distinct yet related thermal characteristics of Heating and Cooling Loads. Positive and negative coefficients across the terms indicate varying contributions of specific input combinations, with some attributes exerting stronger influence on heating efficiency while others affect cooling performance. Overall, the symbolic representation demonstrates the model’s ability to uncover interpretable and physically meaningful relationships between building parameters and energy demands.

Moreover, the symbolic functions enable the extraction of correlations between inputs and outputs, which can facilitate feature engineering. Figure~\ref{fig:enb_c} illustrates an example of correlation analysis between the input and output variables in the symbolic functions. The nonlinear relationships observed across most inputs indicate varying degrees of influence on heating performance. Attributes such as $x_3$, $x_4$, and $x_5$ exhibit strong positive nonlinear trends, suggesting their significant contribution to heating efficiency, whereas $x_1$ and $x_2$ show inverse relationships at lower ranges. In contrast, $x_6$, $x_7$, and $x_8$demonstrate minimal variation, implying limited impact on the output. These insights can guide feature engineering by prioritizing highly correlated and nonlinear attributes while reducing the dimensionality through the exclusion or transformation of less influential features.

\begin{figure*}[htbp]
    \centering
    \includegraphics[width=\linewidth]{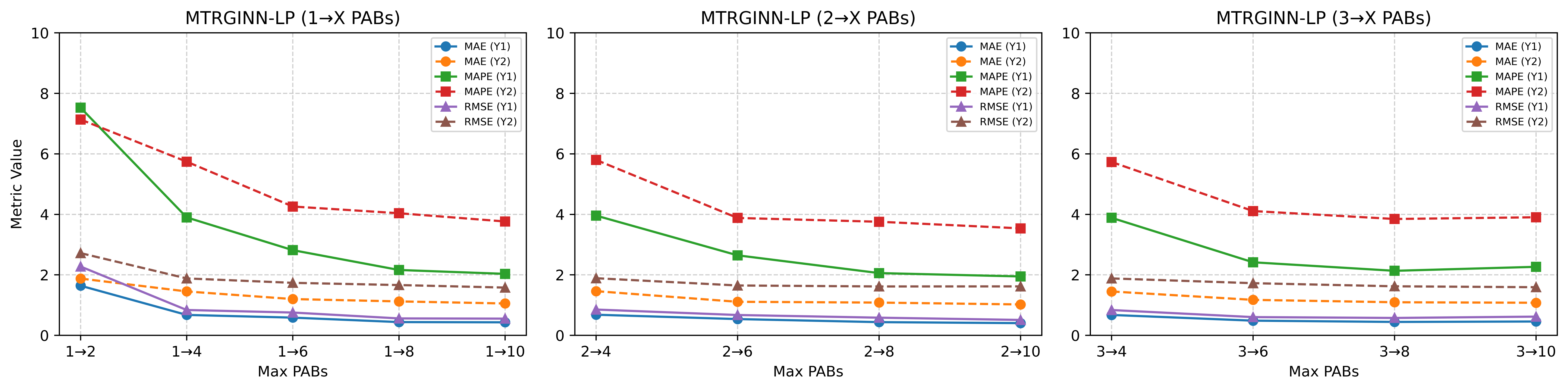}

    \caption{Hyperparameter effects on performance for the Energy Efficiency datasets. It examines how different configurations of the maximum and initial numbers of PABs influence the multi-task regression performance in terms of MAE, MAPE, and RMSE. The initial number of PABs is set to 1, 2, or 3, while the maximum number of PABs is set to 2, 4, 6, 8, and 10.}
    \label{fig:enb_r}
\end{figure*}

\begin{table*}[h!]
\centering
\setlength{\tabcolsep}{2pt} 
\caption{Performance comparison on predicting two targets: Sustainability Score ($Y_1$) and Consumer Trend Index ($Y_2$), on the Sustainable Agriculture dataset.}
\begin{tabular}{lccccccccc}
\hline
\textbf{Model} & \textbf{MAE (Y1)} & \textbf{MAE (Y2)} & \textbf{Avg. MAE} & \textbf{MAPE (Y1)} & \textbf{MAPE (Y2)} & \textbf{Avg. MAPE} & \textbf{RMSE (Y1)} & \textbf{RMSE (Y2)} & \textbf{Avg. RMSE} \\
\hline
Linear Model   			& 0.64 & 0.25 & 0.44  & 26.31 & 5.72 & 16.01  & 0.84 & 0.29 & 0.56  \\
MLP            			& 0.05 & 0.13 & 0.09  & 1.70 & 3.01 & 2.35  & 0.06 & 0.17 & 0.11  \\
Random Forest  		& 0.54 & 0.19 & 0.36  & 22.16 & 4.33 & 13.24  & 0.72 & 0.22 & 0.47  \\
GP             			& 0.65 & 0.26 & 0.45 & 26.38 & 5.91 & 16.14 & 0.84 & 0.30 & 0.57  \\
SVR             			& 0.34 & 0.12 & 0.23  & 18.62 & 2.78 & 10.45  & 0.66 & 0.15 & 0.40  \\

\hline
\hline
\textbf{MTRGINN-LP}             	     	& 0.45 & 0.21 & 0.33  & 17.43 & 4.91 & 11.17 & 0.65 & 0.26 & 0.45 \\
\textbf{MTRGINN-LP (Eq.)}             	& 0.56 & 0.26 & 0.41  & 20.59 & 5.99 & 13.29 & 0.75 & 0.32 & 0.53  \\
\hline
\end{tabular}
\label{tab-arg-1}
\end{table*}

Additionally,  Figure~\ref{fig:enb_r} illustrates the effects of hyperparameter configurations—specifically, the number of initial and maximum PABs—on the performance of the proposed model for the Energy Efficiency datasets. Across all settings, increasing the maximum number of PABs consistently improves performance, as reflected by decreasing MAE, MAPE, and RMSE values. This trend indicates that larger PAB capacities enhance the model’s representational ability to capture complex relationships among inputs. Moreover, the performance gains become less pronounced beyond a certain point (e.g., after 8 max blocks), suggesting diminishing returns with excessive model expansion.

When comparing different initial PAB configurations, models initialized with fewer PABs generally achieve lower errors after expansion than those starting with higher initial counts. This implies that beginning with a simpler structure allows for more effective adaptive growth during training, resulting in better generalization. Overall, the results highlight that a moderate initial configuration combined with a sufficiently large maximum capacity (around 8 PABs) yields the best trade-off between model complexity and predictive accuracy. These findings provide valuable guidance for tuning hyperparameters to optimize multi-task regression performance.

\begin{figure*}[htbp]
    \centering
    \includegraphics[width=\linewidth]{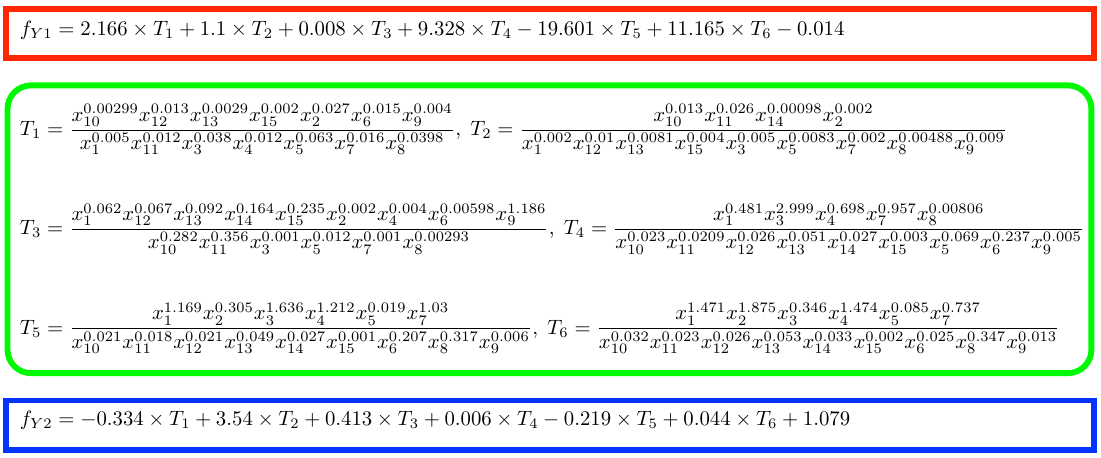}

    \caption{Symbolic functions learned by the proposal model for the Sustainable Agriculture datasets. For each target, the function $f(\cdot)$ consists of  six terms ($T_1$ to $T_6$) and one bias term, corresponding to the optimal number 6 of PABs for the proposed model, where $f_{Y_1}$ and $f_{Y_2}$ are for estimating two targets: Sustainability Score and Consumer Trend Index, respectively. Each term is a combination of fifteen input attributes ($x_1$ to $x_{15}$) with exponential form.}
    \label{fig:agr1}
\end{figure*}

\subsubsection{Sustainable Agriculture}

Table~\ref{tab-arg-1} shows that traditional models such as Linear Regression, GP, and Random Forest perform moderately on the Sustainable Agriculture dataset, with Avg. MAE values ranging from 0.36 (Random Forest) to 0.45 (GP and Linear Model). MLP and SVR achieve lower errors, particularly SVR, which attains an Avg. MAE of 0.23 and Avg. RMSE of 0.40, indicating strong predictive performance on both targets. Against these baselines, MTRGINN-LP demonstrates competitive performance, with an Avg. MAE of 0.33 and Avg. RMSE of 0.45, outperforming Linear Regression, GP, and Random Forest while closely matching the top-performing SVR in some metrics. This indicates that the proposed method effectively leverages correlations between Y1 and Y2 to improve overall accuracy.

The symbolic function, MTRGINN-LP (Eq.), shows slightly higher errors (Avg. MAE of 0.41 and Avg. RMSE of 0.53) than the full MTRGINN-LP model but still surpasses several traditional baselines such as Linear Regression and GP. This demonstrates that the extracted symbolic functions retain meaningful predictive capability while providing interpretable representations. Overall, the results highlight the utility of the MTRGINN-LP framework for achieving a balance between predictive accuracy and model interpretability in sustainable agriculture tasks.

Figure~\ref{fig:agr1} presents the symbolic functions learned by the proposal model for the Sustainable Agriculture datasets. The symbolic functions allows the model to capture intricate, non-linear dependencies between agricultural input factors and sustainability outcomes, suggesting a data-driven discovery of interpretable equations. The coefficients in the linear combinations $f_{Y_1}$ and $f_{Y_2}$ for show that different terms contribute with varying magnitudes and directions to each target. For instance, $f_{Y_1}$ is strongly influenced by $T_4$  (positively) and $T_5$ (negatively), while  $f_{Y_2}$ gives higher weight to $T_2$  and smaller positive or negative contributions from other terms. This indicates that the model can differentiate between the underlying factors that drive sustainability performance and consumer perception. 

In addition, variables $x_{10}$, $x_{12}$, $x_{13}$, and $x_{14}$ emerge as the most influential inputs, appearing repeatedly across multiple terms ($T_1$ to $T_6$) and driving both sustainability and consumer trend predictions through complex, non-linear interactions. Variables such as $x_{3}$, $x_{6}$, $x_{8}$, and $x_{9}$ show moderate influence, fine-tuning the model’s response via small exponential effects, while $x_{1}$, $x_{2}$, $x_{5}$, and $x_{15}$ appear infrequently, indicating a limited or contextual role. The presence of multiplicative and exponential combinations highlights strong interdependencies among agricultural factors, suggesting that sustainability outcomes depend on balanced interactions rather than single-variable effects. Moreover, the differing coefficients between $f_{Y_1}$ and $f_{Y_2}$ reveal that the same variables can impact Sustainability Score and Consumer Trend Index in distinct ways, underscoring how consumer perception and ecological performance are related but not identical dimensions of agricultural sustainability.

\begin{figure*}[htbp]
    \centering
    \includegraphics[width=\linewidth]{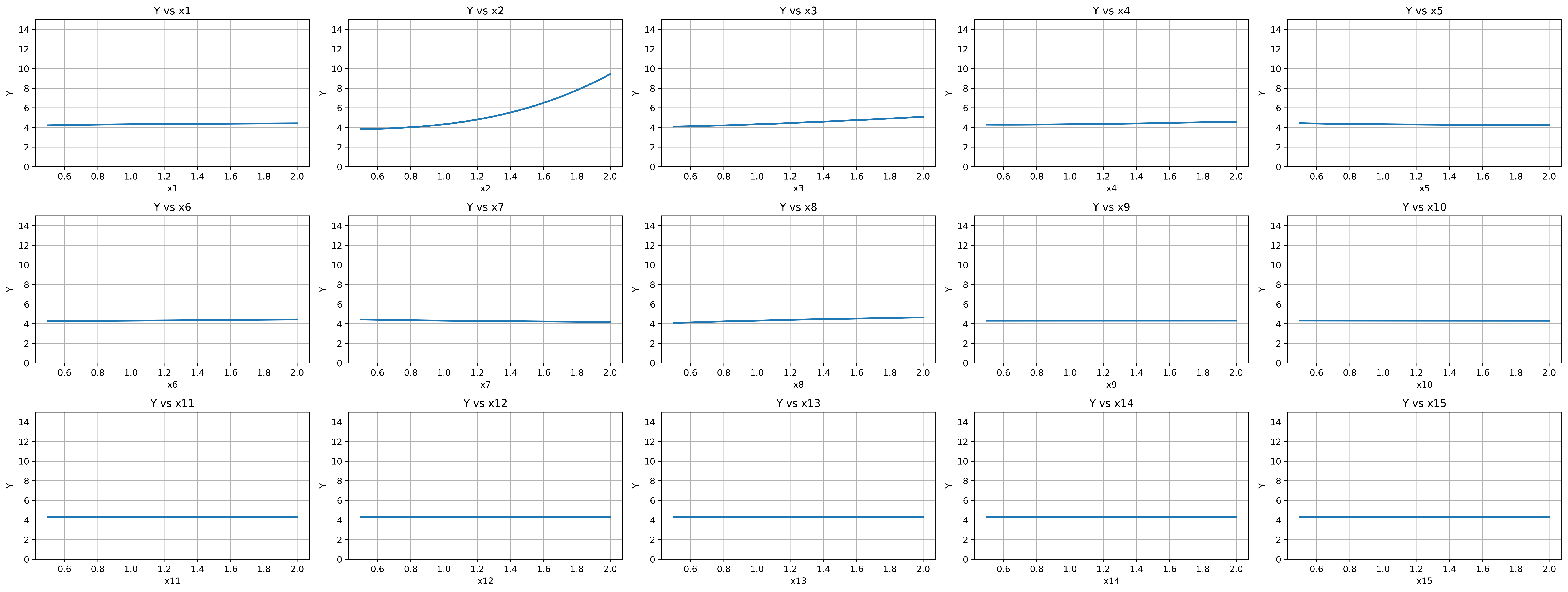}

    \caption{An example of correlation analysis between inputs and outputs in the symbolic functions for the Sustainable Agriculture datasets. Similarly, it is assumed that the input attributes are independent of one another, and the output includes only the target Sustainability Score denoted as $Y$.}
    \label{fig:agr1_r}
\end{figure*}

\begin{figure*}[htbp]
    \centering
    \includegraphics[width=\linewidth]{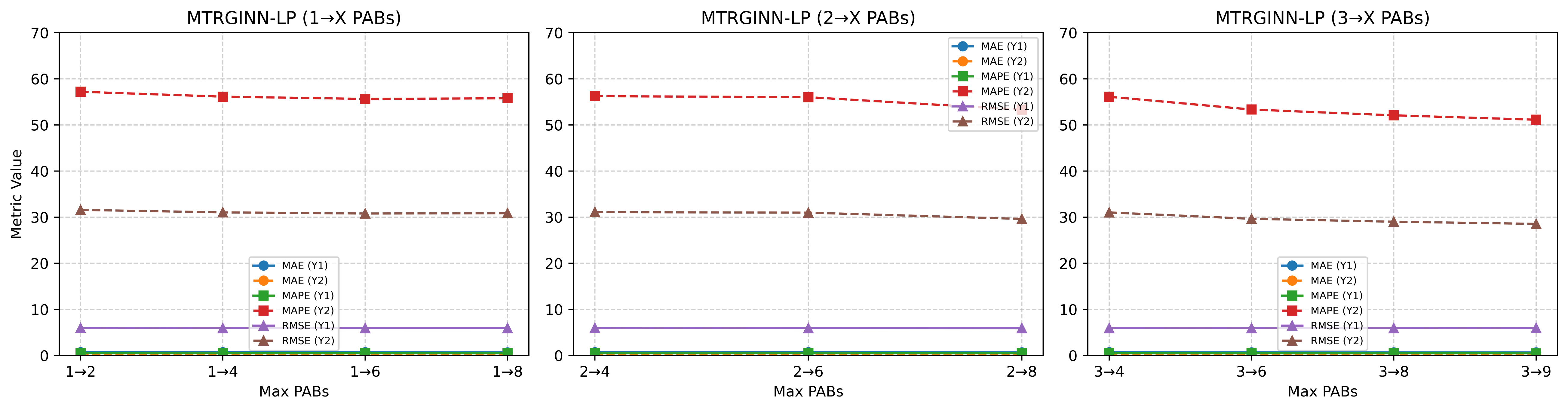}

    \caption{Hyperparameter effects on performance for the Sustainable Agriculture datasets.}
    \label{fig:agr1_h}
\end{figure*}

Figure~\ref{fig:agr1_r} illustrate how each input variable individually influences the target output (Sustainability Score). Most variables exhibit nearly flat relationships with the target, indicating weak or negligible direct correlations. This suggests that no single variable alone strongly determines sustainability performance; rather, sustainability outcomes emerge from complex, multi-variable interactions. A few variables, particularly $x_2$ and $x_3$ , show mild upward trends, implying a modest positive influence on sustainability when these features increase. The general absence of strong linear patterns also supports that the system’s behavior is primarily governed by non-linear and multiplicative relationships among variables rather than by simple additive effects.

From a feature engineering perspective, these observations highlight the importance of constructing composite or interaction-based features to capture the true underlying dynamics of agricultural sustainability. Since most raw inputs lack strong independent predictive power, transforming them through non-linear combinations, ratios, or exponential relationships can significantly enhance model expressiveness. 

Figure~\ref{fig:agr1_h} shows how varying the number of PABs influences model performance across different configurations of the proposed method for the Sustainable Agriculture datasets. As the number of blocks increases in each setup , all performance metrics generally decrease, indicating improved accuracy and model stability. The reduction is particularly evident for MAPE and RMSE, especially for the second target , suggesting that deeper architectures enhance the model’s capacity to capture complex non-linear relationships within the data. However, after a certain point (beyond six or eight blocks), the performance gains plateau, implying diminishing returns from additional network depth. Overall, the results highlight the importance of hyperparameter tuning in balancing model complexity and interpretability, ensuring that the proposed method remains both efficient and robust for sustainability prediction tasks.

\subsubsection{Ablation Study}

\begin{table*}[h!]
\centering
\setlength{\tabcolsep}{2pt} 
\caption{Performance comparison between with and without symbolic loss (SL).}
\begin{tabular}{lccccccccc}

\hline
\textbf{Model} & \textbf{MAE (Y1)} & \textbf{MAE (Y2)} & \textbf{Avg. MAE} & \textbf{MAPE (Y1)} & \textbf{MAPE (Y2)} & \textbf{Avg. MAPE} & \textbf{RMSE (Y1)} & \textbf{RMSE (Y2)} & \textbf{Avg. RMSE} \\
\hline
\multicolumn{10}{c}{Energy Efficiency Datasets} \\
\hline
MTRGINN-LP             	     			& 0.40 & 1.02 & 0.71  & 1.94 & 3.53 & 2.74  & 0.50 & 1.61 & 1.06  \\
MTRGINN-LP w/o SL           	     		&  0.42 & 1.06 & 0.74  & 2.20 & 3.62 & 2.91 & 0.53 & 1.65 & 1.09 \\
MTRGINN-LP (Eq.)             			& 0.56 & 0.26 & 0.41  & 20.59 & 5.99 & 13.29 & 0.75 & 0.32 & 0.53  \\
MTRGINN-LP (Eq.)  w/o SL             	& 33.92 & 27.71 & 30.81  & 271.92 & 172.84 & 222.28 & 52.08 & 41.42 & 46.75  \\
\hline
\multicolumn{10}{c}{Sustainable Agriculture Datasets} \\
\hline
MTRGINN-LP             	     			& 0.45 & 0.21 & 0.33  & 17.43 & 4.91 & 11.17 & 0.65 & 0.26 & 0.45 \\
MTRGINN-LP w/o SL           	     		& 0.47 & 0.22 & 0.34  & 18.40 & 4.93 & 11.66 & 0.67 & 0.26 & 0.46 \\
MTRGINN-LP (Eq.)             			& 0.56 & 0.26 & 0.41  & 20.59 & 5.99 & 13.29 & 0.75 & 0.32 & 0.53 \\
MTRGINN-LP (Eq.)  w/o SL             	& 2.28 & 0.29 & 1.28  & 77.84 & 6.59 & 42.22 & 13.60 & 0.40 & 7.00  \\
\hline
\end{tabular}
\label{tab-abl}
\end{table*}

The ablation study demonstrates the impact of including symbolic loss (SL) in training. As shown in Table~\ref{tab-abl}, the full MTRGINN-LP models consistently outperform their counterparts without SL across all datasets, though the differences are more pronounced in the symbolic variants (MTRGINN-LP (Eq.)). For example, on the Energy Efficiency dataset, MTRGINN-LP achieves an Avg. MAE of 0.71, which slightly increases to 0.74 when SL is removed, indicating a modest benefit for the neural model itself. In contrast, MTRGINN-LP (Eq.) without SL suffers a dramatic performance drop (Avg. MAE from 0.41 to 30.81), highlighting that the symbolic expressions rely heavily on SL to maintain predictive accuracy. Similar trends are observed for the Sustainable Agriculture cases, where the neural models degrade only slightly without SL, but the symbolic variants fail almost completely without it.

These results suggest that the symbolic loss is crucial for extracting meaningful and accurate symbolic representations from the trained proposed method. While the neural MTRGINN-LP is robust enough to retain much of its predictive power even without SL, the interpretability and utility of the extracted symbolic formulas depend strongly on this constraint. Overall, the ablation study confirms that SL effectively guides the learning process to produce stable and interpretable symbolic equations without significantly compromising neural model performance, reinforcing the importance of combining predictive accuracy with symbolic regularization in multi-target regression tasks.

\subsection{Limitations}

Although the proposed method is capable of constructing symbolic expressions for real-world datasets, it still exhibits two key limitations that affect its overall performance and generalization. First, its representational ability is limited because it relies exclusively on exponential forms of inputs to build symbolic functions. While exponential terms can capture certain types of nonlinear relationships, they are insufficient for modeling a broader range of functional behaviors. Incorporating additional function types, such as linear, polynomial, sine, cosine, or logarithmic functions, could significantly enhance the representation ability and adaptability of the generated symbolic expressions. Second, the flexibility of the PABs is constrained, as all PABs are constructed uniformly. This uniform structure restricts the method’s ability to capture diverse patterns and interactions within the data, further limiting the richness and performance of the resulting symbolic functions.

%% file: Relatedwork.tex

An AI system is considered interpretable when its internal mechanisms and outputs can be expressed in human-understandable terms without compromising validity~\cite{graziani2023global}. In healthcare, MacDonald \textit{et al.} emphasize the need for transparent and interpretable AI to satisfy strict standards of safety, fairness, and reliability in clinical decision-making and scientific explanation~\cite{macdonald2022interpretable}. Dibaeinia \textit{et al.} advance interpretability in molecular biology by defining gene regulatory relationships and introducing Counterfactual Inference by Machine Learning and Attribution Models (CIMLA), a tool that uses counterfactual inference and attribution methods to detect differences in gene regulatory networks across biological conditions~\cite{dibaeinia2025interpretable}. In materials and energy research, Peng \textit{et al.} present an interpretable approach based on a generalized additive model with interactive features (GAM-IFI) to predict early-stage battery capacities and explain how coating features influence performance~\cite{peng2024coating}. Baek \textit{et al.} propose harmonic loss as an interpretable alternative to cross-entropy, replacing SoftMax with a scale-invariant HarMax function and computing logits via Euclidean distance; this design enhances interpretability and accelerates convergence by aligning each class with a finite, well-defined center~\cite{baek2025harmonic}. They further develop a multimodal interpretation framework incorporating predictive models for static and time-series data, using perturbation-based global feature importance, permutation importance (PIMP), and SHAP to identify key risk factors and strategies for fall prevention at home~\cite{baek2025interpretable}.

Neuro-Symbolic AI is a hybrid framework that integrates symbolic reasoning with neural and probabilistic methods to improve interpretability, robustness, trustworthiness, and data efficiency for AI systems~\cite{colelough2025neuro,wan2024towards}. Advancements in the field span five core areas: knowledge representation, learning and inference, explainability and trustworthiness, logic and reasoning, and meta-cognition, with rapid growth since 2020 particularly in learning and inference. Recent work includes LogiCity, a customizable first-order-logic-based simulator enabling long-horizon reasoning and complex multi-agent interactions in urban environments~\cite{li2024logicity}; Hagos \textit{et al.}’s analysis of Neuro-Symbolic AI’s potential to enhance military decision-making, automate intelligence workflows, and strengthen autonomous systems~\cite{hagos2024neuro}; and Zhang \textit{et al.}’s four-level framework for representation spaces, covering five space types, five information modalities, symbolic logic methods, and three neuro–symbolic collaboration strategies~\cite{zhang2024bridging}. Despite these advances, substantial gaps remain in explainability, trustworthiness, and meta-cognition, requiring interdisciplinary efforts to achieve more reliable and context-aware AI systems~\cite{colelough2025neuro}. Promising future directions include developing unified representation spaces to reduce information loss and improve representational efficiency~\cite{zhang2024neuro}, and enhancing large language models by embedding them within broader neuro-symbolic reasoning frameworks~\cite{hammond2023large}.

%% file: Conclusion.tex

Symbolic Regression (SR) has shown great potential in XAI by generating interpretable mathematical expressions on the data. Despite its promise, SR has been limited by its focus on well-understood scientific datasets and single-output regression, which restricts its applicability to real-world problems involving multiple interdependent outputs. In this study, we addressed these limitations by proposing MTRGINN-LP, an interpretable neural network for multi-target symbolic regression. Experimental results on practical multi-target applications demonstrate that MTRGINN-LP achieves competitive predictive performance without compromising explainability. These results highlight its potential for extending symbolic regression to more complex, real-world multi-output tasks, bridging the gap between interpretability and practical applicability.

In the future,  there are two key directions to enhance the proposed method. First, the representational capacity of the symbolic functions could be improved by extending beyond the current reliance on exponential forms. Incorporating a wider variety of function types—such as linear, polynomial, sine, cosine, or logarithmic terms—would enable the model to capture a broader range of nonlinear relationships and complex functional behaviors, improving both adaptability and accuracy. Second, instead of constructing all PABs uniformly, exploring heterogeneous or dynamically adaptive PAB architectures could allow the model to better capture diverse patterns and interactions within the data.

%% file: main.bbl
\begin{thebibliography}{10}
\providecommand{\url}[1]{#1}
\csname url@samestyle\endcsname
\providecommand{\newblock}{\relax}
\providecommand{\bibinfo}[2]{#2}
\providecommand{\BIBentrySTDinterwordspacing}{\spaceskip=0pt\relax}
\providecommand{\BIBentryALTinterwordstretchfactor}{4}
\providecommand{\BIBentryALTinterwordspacing}{\spaceskip=\fontdimen2\font plus
\BIBentryALTinterwordstretchfactor\fontdimen3\font minus
  \fontdimen4\font\relax}
\providecommand{\BIBforeignlanguage}[2]{{%
\expandafter\ifx\csname l@#1\endcsname\relax
\typeout{** WARNING: IEEEtran.bst: No hyphenation pattern has been}%
\typeout{** loaded for the language `#1'. Using the pattern for}%
\typeout{** the default language instead.}%
\else
\language=\csname l@#1\endcsname
\fi
#2}}
\providecommand{\BIBdecl}{\relax}
\BIBdecl

\bibitem{chai2021deep}
J.~Chai, H.~Zeng, A.~Li, and E.~W. Ngai, ``Deep learning in computer vision: A
  critical review of emerging techniques and application scenarios,'' Machine
  Learning with Applications, vol.~6, 2021, p. 100134.

\bibitem{pouyanfar2018survey}
S.~Pouyanfar, S.~Sadiq, Y.~Yan, H.~Tian, Y.~Tao, M.~P. Reyes, M.-L. Shyu, S.-C.
  Chen, and S.~S. Iyengar, ``A survey on deep learning: Algorithms, techniques,
  and applications,'' ACM computing surveys (CSUR), vol.~51, no.~5, 2018, pp.
  1--36.

\bibitem{ji2025comprehensive}
Y.~Ji, Y.~Sun, Y.~Zhang, Z.~Wang, Y.~Zhuang, Z.~Gong, D.~Shen, C.~Qin, H.~Zhu,
  and H.~Xiong, ``A comprehensive survey on self-interpretable neural
  networks,'' CoRR, 2025.

\bibitem{zhang2024neuro}
X.~Zhang and V.~S. Sheng, ``Neuro-symbolic ai: Explainability, challenges, and
  future trends,'' arXiv preprint arXiv:2411.04383, 2024.

\bibitem{guidotti2018survey}
R.~Guidotti, A.~Monreale, S.~Ruggieri, F.~Turini, F.~Giannotti, and
  D.~Pedreschi, ``A survey of methods for explaining black box models,'' ACM
  computing surveys (CSUR), vol.~51, no.~5, 2018, pp. 1--42.

\bibitem{ribeiro2016should}
M.~T. Ribeiro, S.~Singh, and C.~Guestrin, ``" why should i trust you?"
  explaining the predictions of any classifier,'' in Proceedings of the 22nd
  ACM SIGKDD international conference on knowledge discovery and data mining,
  2016, pp. 1135--1144.

\bibitem{lundberg2017unified}
S.~M. Lundberg and S.-I. Lee, ``A unified approach to interpreting model
  predictions,'' Advances in neural information processing systems, vol.~30,
  2017.

\bibitem{somvanshi2025bridging}
S.~SOMVANSHI, M.~M. ISLAM, A.~RAFE, A.~G. TUSTI, A.~CHAKRABORTY, A.~BAITULLAH,
  T.~I. CHOWDHURY, N.~ALNAWMASI, A.~DUTTA, and S.~DAS, ``Bridging the black
  box: A survey on mechanistic interpretability in ai,'' J. ACM, vol.~37,
  no.~4, 2025.

\bibitem{makke2024interpretable}
N.~Makke and S.~Chawla, ``Interpretable scientific discovery with symbolic
  regression: a review,'' Artificial Intelligence Review, vol.~57, no.~1, 2024,
  p.~2.

\bibitem{koza1994genetic}
J.~R. Koza, ``Genetic programming as a means for programming computers by
  natural selection,'' Statistics and computing, vol.~4, no.~2, 1994, pp.
  87--112.

\bibitem{wang2019symbolic}
Y.~Wang, N.~Wagner, and J.~M. Rondinelli, ``Symbolic regression in materials
  science,'' MRS communications, vol.~9, no.~3, 2019, pp. 793--805.

\bibitem{angelis2023artificial}
D.~Angelis, F.~Sofos, and T.~E. Karakasidis, ``Artificial intelligence in
  physical sciences: Symbolic regression trends and perspectives,'' Archives of
  Computational Methods in Engineering, 2023, p.~1.

\bibitem{dong2025recent}
J.~Dong and J.~Zhong, ``Recent advances in symbolic regression,'' ACM Computing
  Surveys, vol.~57, no.~11, 2025, pp. 1--37.

\bibitem{aldeia2025call}
G.~S.~I. Aldeia, H.~Zhang, G.~Bomarito, M.~Cranmer, A.~Fonseca, B.~Burlacu,
  W.~G. La~Cava, and F.~O. de~Fran{\c{c}}a, ``Call for action: towards the next
  generation of symbolic regression benchmark,'' arXiv preprint
  arXiv:2505.03977, 2025.

\bibitem{ranasinghe2024ginn}
N.~Ranasinghe, D.~Senanayake, S.~Seneviratne, M.~Premaratne, and S.~Halgamuge,
  ``Ginn-lp: A growing interpretable neural network for discovering
  multivariate laurent polynomial equations,'' in Proceedings of the AAAI
  Conference on Artificial Intelligence, vol.~38, no.~13, 2024, pp.
  14\,776--14\,784.

\bibitem{crawshaw2020multi}
M.~Crawshaw, ``Multi-task learning with deep neural networks: A survey,'' arXiv
  preprint arXiv:2009.09796, 2020.

\bibitem{zhang2021survey}
Y.~Zhang and Q.~Yang, ``A survey on multi-task learning,'' IEEE transactions on
  knowledge and data engineering, vol.~34, no.~12, 2021, pp. 5586--5609.

\bibitem{deng2013new}
L.~Deng, G.~Hinton, and B.~Kingsbury, ``New types of deep neural network
  learning for speech recognition and related applications: An overview,'' in
  2013 IEEE international conference on acoustics, speech and signal
  processing.\hskip 1em plus 0.5em minus 0.4em\relax IEEE, 2013, pp.
  8599--8603.

\bibitem{ramsundar2015massively}
B.~Ramsundar, S.~Kearnes, P.~Riley, D.~Webster, D.~Konerding, and V.~Pande,
  ``Massively multitask networks for drug discovery,'' arXiv preprint
  arXiv:1502.02072, 2015.

\bibitem{girshick2015fast}
R.~Girshick, ``Fast r-cnn,'' in Proceedings of the IEEE international
  conference on computer vision, 2015, pp. 1440--1448.

\bibitem{collobert2008unified}
R.~Collobert and J.~Weston, ``A unified architecture for natural language
  processing: Deep neural networks with multitask learning,'' in Proceedings of
  the 25th international conference on Machine learning, 2008, pp. 160--167.

\bibitem{ruder2017overview}
S.~Ruder, ``An overview of multi-task learning in deep neural networks,'' arXiv
  preprint arXiv:1706.05098, 2017.

\bibitem{graziani2023global}
M.~Graziani, L.~Dutkiewicz, D.~Calvaresi, J.~P. Amorim, K.~Yordanova, M.~Vered,
  R.~Nair, P.~H. Abreu, T.~Blanke, V.~Pulignano et~al., ``A global taxonomy of
  interpretable ai: unifying the terminology for the technical and social
  sciences,'' Artificial intelligence review, vol.~56, no.~4, 2023, pp.
  3473--3504.

\bibitem{macdonald2022interpretable}
S.~MacDonald, K.~Steven, and M.~Trzaskowski, ``Interpretable ai in healthcare:
  Enhancing fairness, safety, and trust,'' in Artificial Intelligence in
  Medicine: Applications, Limitations and Future Directions.\hskip 1em plus
  0.5em minus 0.4em\relax Springer, 2022, pp. 241--258.

\bibitem{dibaeinia2025interpretable}
P.~Dibaeinia, A.~Ojha, and S.~Sinha, ``Interpretable ai for inference of causal
  molecular relationships from omics data,'' Science Advances, vol.~11, no.~7,
  2025, p. eadk0837.

\bibitem{peng2024coating}
Q.~Peng, Y.~Liu, Y.~Jin, X.-G. Yang, R.~Wang, and K.~Liu, ``Coating feature
  analysis and capacity prediction for digitalization of battery manufacturing:
  An interpretable ai solution,'' IEEE Transactions on Systems, Man, and
  Cybernetics: Systems, 2024.

\bibitem{baek2025harmonic}
D.~D. Baek, Z.~Liu, R.~Tyagi, and M.~Tegmark, ``Harmonic loss trains
  interpretable ai models,'' arXiv preprint arXiv:2502.01628, 2025.

\bibitem{baek2025interpretable}
J.~Baek, Y.~Li, L.~Lim, and J.~W. Chong, ``An interpretable ai for smart homes:
  Identifying fall prevention strategies for older adults using multimodal deep
  learning,'' IEEE Journal of Biomedical and Health Informatics, 2025.

\bibitem{colelough2025neuro}
B.~C. Colelough and W.~Regli, ``Neuro-symbolic ai in 2024: A systematic
  review,'' arXiv preprint arXiv:2501.05435, 2025.

\bibitem{wan2024towards}
Z.~Wan, C.-K. Liu, H.~Yang, C.~Li, H.~You, Y.~Fu, C.~Wan, T.~Krishna, Y.~Lin,
  and A.~Raychowdhury, ``Towards cognitive ai systems: a survey and prospective
  on neuro-symbolic ai,'' arXiv preprint arXiv:2401.01040, 2024.

\bibitem{li2024logicity}
B.~Li, Z.~Li, Q.~Du, J.~Luo, W.~Wang, Y.~Xie, S.~Stepputtis, C.~Wang,
  K.~Sycara, P.~Ravikumar et~al., ``Logicity: Advancing neuro-symbolic ai with
  abstract urban simulation,'' Advances in Neural Information Processing
  Systems, vol.~37, 2024, pp. 69\,840--69\,864.

\bibitem{hagos2024neuro}
D.~H. Hagos and D.~B. Rawat, ``Neuro-symbolic ai for military applications,''
  IEEE Transactions on Artificial Intelligence, 2024.

\bibitem{zhang2024bridging}
X.~Zhang and V.~S. Sheng, ``Bridging the gap: representation spaces in
  neuro-symbolic ai,'' arXiv preprint arXiv:2411.04393, 2024.

\bibitem{hammond2023large}
K.~J. Hammond and D.~B. Leake, ``Large language models need symbolic ai.'' in
  NeSy, 2023, pp. 204--209.

\end{thebibliography}
